\documentclass[10pt,twocolumn,letterpaper]{article}

\usepackage[pagenumbers]{cvpr} % To force page numbers, e.g. for an arXiv version

\usepackage{graphicx}
\usepackage{amsmath}
\usepackage{amssymb}
\usepackage{booktabs}

\usepackage[pagebackref,breaklinks,colorlinks]{hyperref}

\usepackage[capitalize]{cleveref}
\crefname{section}{Sec.}{Secs.}
\Crefname{section}{Section}{Sections}
\Crefname{table}{Table}{Tables}
\crefname{table}{Tab.}{Tabs.}

\usepackage{xspace}

\newcommand*{\datasetname}{Seal Watch\@\xspace}
\usepackage{multirow}

\def\cvprPaperID{5} % *** Enter the CVPR Paper ID here
\def\confName{CVPR}
\def\confYear{2022}

\begin{document}

\title{Fine-Grained Counting with Crowd-Sourced Supervision}

\author{Justin Kay\\
Ai.Fish\\
% {\tt\small justin@ai.fish}
\and
Catherine M. Foley\\
Stony Brook University\\
\and
Tom Hart\\
University of Oxford\\
}
\maketitle

%%%%%%%%% ABSTRACT
\begin{abstract}

Crowd-sourcing is an increasingly popular tool for image analysis in animal ecology. Computer vision methods that can utilize crowd-sourced annotations can help scale up analysis further. In this work we study the potential to do so on the challenging task of fine-grained counting. As opposed to the standard crowd counting task, fine-grained counting also involves classifying attributes of individuals in dense crowds. We introduce a new dataset from animal ecology to enable this study that contains 1.7M crowd-sourced annotations of 8 fine-grained classes. It is the largest available dataset for fine-grained counting and the first to enable the study of the task with crowd-sourced annotations. We introduce methods for generating aggregate ``ground truths'' from the collected annotations, as well as a counting method that can utilize the aggregate information. Our method improves results by 8\% over a comparable baseline, indicating the potential for algorithms to learn fine-grained counting using crowd-sourced supervision.

\end{abstract}

%%%%%%%%% BODY TEXT
\section{Introduction}
\label{sec:intro}

Automated image capture technologies have enabled large-scale, non-invasive, and long-term observation of animal species in the wild, however the large quantities of imagery produced can overwhelm manual analysis capabilities.  Ecologists have increasingly utilized citizen science platforms such as Zooniverse~\cite{simpson2014zooniverse} and iNaturalist~\cite{inaturalist} to crowd-source this analysis to larger groups of paid or volunteer workers. Using these crowd-sourced annotations to train computer vision algorithms offers a way to further scale up analysis without requiring additional human effort, but requires methods that can account for variability and discrepancies in the collected annotations.

\begin{figure}
  \centering
%   \fbox{\rule{0pt}{2.5in} \rule{0.9\linewidth}{0pt}}
   \includegraphics[width=1.\linewidth]{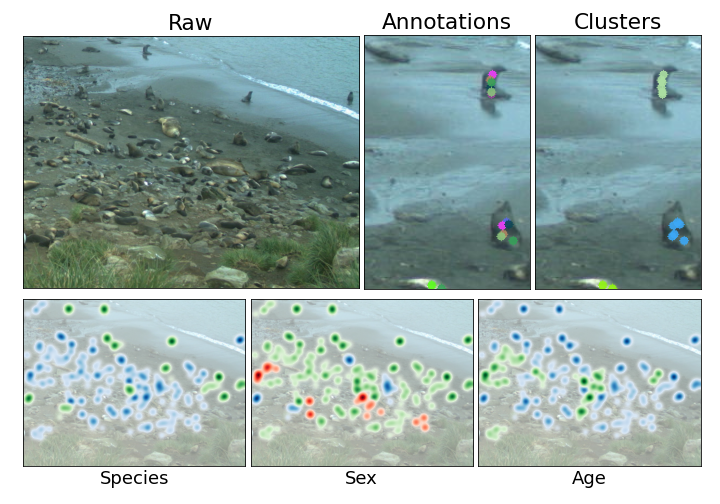}
   \caption{\textbf{Top Left:} Example image from the \datasetname dataset. Seals are difficult to differentiate from rocks, background, and each other. \textbf{Top Middle:} Zoomed example with all crowd-sourced dot annotations overlaid. This image was annotated by 11 users, each shown as a different color. \textbf{Top Right:} Aggregated annotations for each seal after clustering as described in \cref{sec:gt}. Each cluster shown as a different color. \textbf{Bottom:} Fine-grained density maps for each attribute. Binary classifications shown as blue/green and ``unknown'' classifications shown in red.}
   \label{fig:fig1}
\end{figure}

In this work we study the use of crowd-sourced image annotations to train computer vision algorithms on the challenging task of \textit{fine-grained counting}. Recently introduced in \cite{go2021fine,wan2021fine}, fine-grained counting extends crowd counting---estimating the number of individuals in a densely crowded scene, typically used for counting human crowds---to a fine-grained multi-class scenario. While previous work has proposed methods for utilizing crowd-sourced annotations for image classification \cite{kovashka2016crowdsourcing} and single-class counting~\cite{arteta2016counting}, we are the first to study their use for training algorithms for fine-grained counting.

To enable this study we introduce a new image dataset curated from the Seal Watch project~\cite{sealwatch}, an ongoing citizen science effort to collect crowd-sourced observations of seal populations in time-lapse imagery. The goals of the project are to count and classify the species, sex, and age of all visible seals, which we identify as a challenging real-world example of the fine-grained counting task. The dataset includes over 1.68 million crowd-sourced annotations collected from 7,364 volunteers in 5,633 images, making it more than 50\% larger than existing datasets for fine-grained counting and the first to support the study of this task with crowd-sourced annotations.

\begin{figure*}
  \centering
%   \fbox{\rule{0pt}{1in} \rule{0.9\linewidth}{0pt}}
  \includegraphics[width=1.\linewidth]{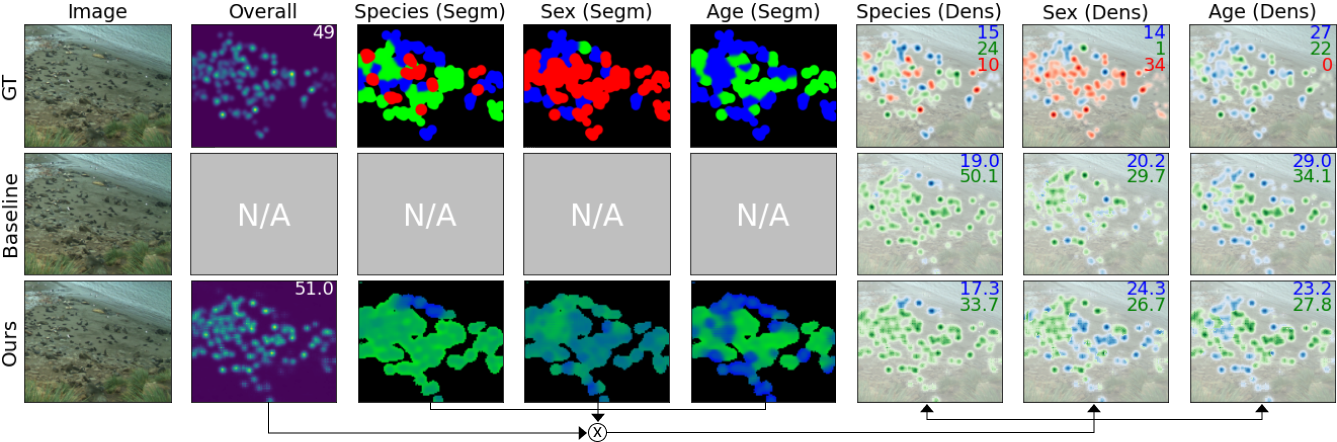}
  \caption{Ground truth (top) and predicted (middle and bottom) density and segmentation maps for an example image. \textbf{From left:} Raw image; Overall (class-agnostic) density map; Segmentation maps for species, sex, and age respectively. Blue/green are soft segmentations for known classes, red is a mask for ``unknown'' classifications, and black is background; Fine-grained density maps for species, sex, and age respectively. The baseline method predicts fine-grained density maps directly, and may produce inconsistent counts between attributes. In our method, predictions are obtained by multiplying the overall density map by each predicted soft segmentation, indicated by arrows. Ground truth and predicted counts are reported. Note that predicted counts may be fractional and do not include an ``unknown'' class.}
  \label{fig:inference_examples}
\end{figure*}

Additionally, we introduce a method for processing these annotations into aggregated ground truths, and propose novel extensions to a popular crowd-counting approach that can make use of the aggregate information. Our approach follows the density-estimation counting paradigm, whereby we predict per-pixel object densities over entire images and integrate densities to obtain overall counts. We harness the spatial variability in the crowd-sourced annotations to generate ground-truth density maps that encode targets' spatial extent and propose a method for loss masking that accounts for regions of difficult classification. We also improve results by extending the counting network with a segmentation branch to predict overall counts and fine-grained classifications in parallel. We show that our approach improves results over a comparable baseline.

\begin{table}
  \centering
  \begin{tabular}{l|r|r|r|r}
    \hline\noalign{\smallskip}
    Dataset & \# Cls & \# Img & \# Obj & \# Anno \\
    \noalign{\smallskip}
    \hline\hline
    \noalign{\smallskip}
    Wan \etal \cite{wan2021fine} & 2 & 3,728 & 112k & 112k  \\
    KR-GRUIDAE \cite{go2021fine} & 5 & 1,423 & 31k & 62k  \\
    \midrule
    \datasetname & \textbf{8} & \textbf{5,633} & \textbf{192k} & \textbf{1.7M} \\
    \hline
    \end{tabular}
  \caption{Datasets for fine-grained counting. \datasetname includes more classes, images, object instances, and annotations than existing options.}
  \label{tab:datasets}
\end{table}
%1.6x more classes, 1.5x more images, 1.7x more object instances, and 15x more annotations than existing options.

\section{Related work}
\label{sec:related}

\noindent
\textbf{Crowd counting datasets} 
%Existing 
Image datasets for crowd counting deal primarily with counting people in crowded urban scenes \cite{wang2020nwpu,hsieh2017drone,idrees2018composition,zhang2015cross,zhang2016data,zhang2016single}. Wan \etal~\cite{wan2021fine} extend this task to the multi-class setting by adding fine-grained attributes to existing human counting datasets. Go \etal \cite{go2021fine} introduce KR-GRUIDAE, a fine-grained counting dataset consisting of 5 bird classes. In comparison to existing datasets for fine-grained counting, \datasetname contains more classes, images, object instances, and annotations. See \cref{tab:datasets}.
%Several works introduce datasets for single-class animal counting in the wild~\cite{arteta2016counting,naude_johannes_j_2019_3234780,center2020noaa}.

\noindent
\textbf{Crowd counting methods} We focus on density-based crowd counting approaches due to their prevalence in re-
cent literature \cite{gao2020cnn,lempitsky2010learning,zhang2016single,li2018csrnet}. These methods estimate per-pixel crowd densities and integrate over all pixels to obtain the total count for an image. 
%On the other hand, detection-based approaches \cite{sam2020locate,song2021rethinking} have recently shown competitive results by directly predicting object locations. 
Existing methods primarily target single-class counting and assume a single ground truth location for each individual. In contrast, \datasetname contains 8 classes and up to 94 crowd-sourced ``ground truth'' locations per individual. Our method extends existing approaches to perform multi-class counting and make use of the additional annotations.
% crowd counting approaches due to their prevalence in recent literature \cite{}. These methods aim to predict image-level crowd density maps and then integrate these maps to obtain counts. [Description of methods ; how ground truth is generated] [Method for fine-grained] [What we do differently -- we have no source of ground truth and other methods won't work]

\noindent
\textbf{Learning from crowd-sourced dot annotations} In our work most annotators are anonymous and contribute very few annotations (see \cref{fig:data}A), precluding the use of techniques for crowd-sourced data that create models of each user's annotation quality, \eg~\cite{dawid1979maximum,van2018lean}. Arteta \etal~\cite{arteta2016counting} introduce a method for learning to count from single-class crowd-sourced dot annotations using a segmentation-guided density prediction network, using annotator variability to improve foreground/background segmentation. Jones \etal~\cite{jones2018time} instead cluster nearby annotations into ``consensus clicks'' to reduce the annotation set to one dot per object. Our method can be seen as a hybrid of these two approaches that is extended to the multi-class setting. 
%We cluster annotations belonging to the same object while also making use of spatial information to improve both density and segmentation map prediction.
We cluster annotations of the same object and make use of annotation variability to improve both density and segmentation prediction.

%We focus on learning from crowd-sourced annotations and must account for annotation discrepancies between individual annotators.
%Crowd counting annotations are typically provided as ``dots'', \ie a single $(x,y)$ coordinate for each object center.

\section{Dataset}
\label{sec:dataset}

\begin{figure*}
  \centering
  \includegraphics[width=1.\linewidth]{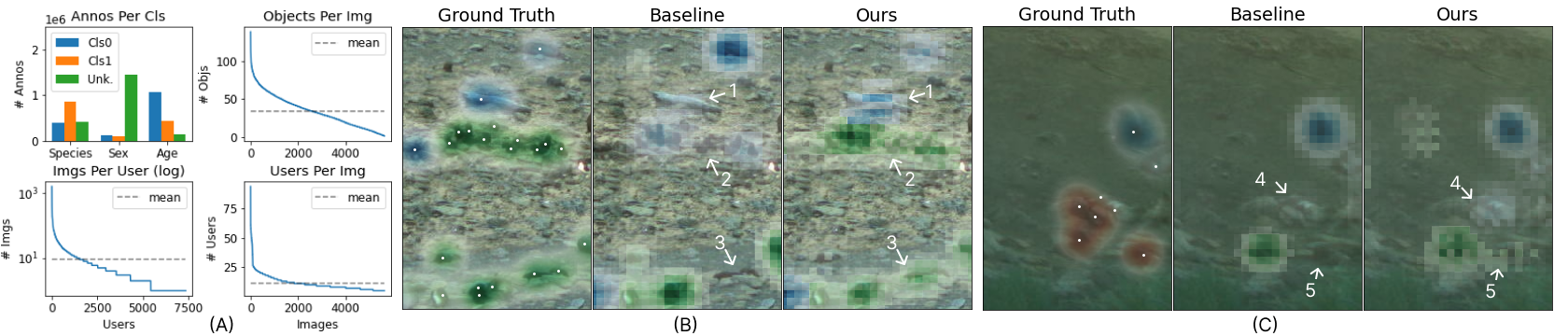}
  \caption{\textbf{(A):} Dataset statistics. Top row: Num. annotations per class and num. objects per image after performing clustering. Bottom row: Num. images annotated per user (log y scale), and num. users per image. Mean values shown in gray. \textbf{(B–C):}~Qualitative prediction comparisons between baseline and our method on the age attribute. Images shown at 5x zoom and ground truth objects marked with white dots for clarity. Prominent differences indicated with arrows. In B, we see our method is better able to distinguish both classes from the background. In C, we also see an improvement in prediction performance in regions with unknown ground truth classifications.}
  \label{fig:data}
  \vspace{-4mm}
\end{figure*}

%\noindent
%\textbf{Images and annotations} 
The \datasetname dataset consists of imagery from a single time-lapse camera deployed in the Elsehul bay in South Georgia for the entirety of the 2014--2015 seal breeding season. 
%The camera captured one image per hour for a period of three months (November 2014--January 2015). 
%See \cref{fig:data}B for the distribution of images and annotations over time. 
%Initial feedback indicated that annotators were not able to perform the counting task using images at their native resolution, so we generated random 1024x768px crops for annotation. Even in these smaller crops the challenges of identifying seals are significant 
% Identifying seals presents significant challenges due to variations in scale, similarity with background environment, and similarities between seals of different classes (see \cref{fig:data}A).
% Each image was annotated by between 5 and 94 users who were asked to place a single dot in the center of each seal. %Because annotators were all volunteers, their level of
Users were asked to place a single dot in the center of each seal.
The number of annotations contributed by each user varied significantly (see \cref{fig:data}A), as did the location on each animal where they placed their dots (\cref{fig:fig1}).
% : the mean number of images annotated by each user is 9 with a standard deviation of 34, a minimum of 1 and a maximum of 1608 (see \cref{fig:data}B). Annotators' dots differ spatially (see \cref{fig:fig1}). 

The 8 classes in \datasetname are broken down into 3 attributes, each consisting of a binary classification: species (elephant/fur), sex (male/female), and age (adult/pup). 
Users could respond ``unknown'' for classifications they could not perform. The distribution of class annotations is shown in \cref{fig:data}A. The sex attribute was the most challenging, indicated by the large number of ``unknown'' responses.

\noindent
\textbf{Data split} We split the data temporally. Our training and validation sets contain data from November--December 2014, and our test set consists of imagery from January 2015. This gives us 4,849 training images, 453 validation images, and 331 test images, with 157,188, 19,432, and 15,102 object instances, respectively.

\section{Methods}
\label{sec:method}

\subsection{Ground truth generation}
\label{sec:gt}

\noindent
\textbf{Dot aggregation} Similar to \cite{jones2018time}, we perform a hierarchical clustering of all users' annotations, enforcing a connectivity constraint that prevents two annotations from the same person from being in the same cluster, and require a minimum cluster size of 2.
For each point cluster we assign a class for each attribute based on majority voting. Clusters with no user responses for a particular attribute are classified as ``unknown'' for that attribute.

\noindent
\textbf{Density map generation} We compare two methods for generating ground truth density maps. The first is the standard fixed-kernel approach introduced in \cite{zhang2016single}. 
We use the medoid of each cluster as ground truth locations and fix the kernel bandwidth $\sigma=12$ based on initial experiments.

We introduce a second method that utilizes all points in each of our generated clusters. 
For each cluster, indexed by $k$, we set the density value at the pixel location of the $j$th point in the cluster, $j \in \{1,...,J_k\}$, to $1/J_k$, and convolve with a 2D Gaussian filter. 
Thus the integrated density for each ground truth object equals 1 as usual, but the clusters allow the density map to reflect the spatial variability in annotator's dot locations. See \cref{fig:fig1}.

\noindent
\textbf{Segmentation map generation} We use the ground truth class density maps to generate soft segmentation maps that enable multi-task training (\cref{sec:dens_methods}) and aid in error calculation in regions with unknown classifications (\cref{sec:eval}). We follow the method proposed in \cite{wan2021fine}, however we calculate separate masks for each attribute. The goal for each segmentation map is to indicate the contribution of each class to the overall count at each pixel with non-zero density. That is, given image dimensions $(w,h)$, the soft segmentation map for the $c$th class of attribute $a$, $S_{a,c} \in \mathbb{R}^{w,h}$ , is:

\begin{equation}
    S_{a,c} = \frac{D_{a,c}}{\sum_{c'=1}^{C_a}D_{a,c'}}
\end{equation}

Where ${D}$ is ground truth density and $C_a=2$ is the number of classes for attribute $a$.
We also add a background segmentation channel by thresholding the overall density maps.

\subsection{Counting approach}
\label{sec:dens_methods}

\noindent
\textbf{Baseline} We choose CSRNet~\cite{li2018csrnet} as our baseline due to its popularity in recent literature. 
We expand the final convolutional layer to predict one density map per class and use a MSE loss $L^c$ to optimize each class density map separately. We add an additional MSE loss for the total object count:

\begin{equation}
    L^t = \sum_{a=1}^{A} MSE(\sum_{c=1}^{C_a} D_{a,c}, \sum_{c=1}^{C_a} \hat{D}_{a,c})
\end{equation}

% Where $\hat{D}_{a,c}$ is the predicted density for the $c$th class of attribute $a$.
Where $\hat{D}$ is predicted density and $A=3$ is the total number of attributes.

\noindent
\textbf{Multi-task network} The baseline predicts independent density maps for each class, thus total counts for each attribute may be inconsistent (see \cref{fig:inference_examples}). To address this we modify the architecture to predict a single density map and add a multi-class segmentation branch to predict fine-grained classifications for each pixel. Final class density maps are obtained by element-wise multiplying these two outputs together. See \cref{fig:inference_examples}. 
We add an additional segmentation loss $L^s$ which is a soft cross entropy loss as in \cite{wan2021fine}.

% We extend the CSRNet architecture with an additional segmentation branch to predict fine-grained classifications for each pixel. The network outputs a single class-agnostic density map as well as a multi-class segmentation map, and then the final class density maps are obtained by element-wise multiplying these two outputs together. See \cref{fig:inference_examples}. 
% We add an additional segmentation loss $L^s$ which is a soft cross entropy loss as in \cite{wan2021fine}.

\noindent
\textbf{Loss masking} To handle unknown classifications, we also mask both $L^c$ and $L^s$ in regions where the predominant class is ``unknown''. 

\subsection{Evaluation}
\label{sec:eval}

For all experiments we report overall class-agnostic counting error as mean average error (MAE).
We also introduce a new metric, category-averaged masked MAE (CMMAE), for evaluating fine-grained counting performance when ground truth classifications may be unknown:

\begin{equation}
    CMMAE = \frac{1}{A} \sum_{a=1}^{A} \frac{1}{C_a}  \sum_{c=1}^{C_a} MMAE_{a,c}
\end{equation}

Where $MMAE_{a,c}$ is the ``masked MAE'' for class $c$ of attribute $a$, \ie the MAE calculated only in regions that are not masked due to the ``unknown'' class as described in \cref{sec:dens_methods}.
Note that this masking does not occur when calculating overall MAE; we still want the network to count objects in the ``unknown'' regions but do not penalize classification performance there.

\section{Results}
\label{sec:results}

\begin{table}
  \centering
%   \small{
      \begin{tabular}{c|c|c|c|c|c}
        \hline\noalign{\smallskip}
        % \multirow{2}{*}{Method} & \multirow{2}{*}{MAE} & CM- & \multirow{2}{*}{Species} & \multirow{2}{*}{Sex} & \multirow{2}{*}{Age} \\
        Method & MAE & CMMAE & Species & Sex & Age \\
        % & & MAE & & & \\
        \noalign{\smallskip}
        \hline\hline
        \noalign{\smallskip}
        \multirow{2}{*}{Baseline} & \multirow{2}{*}{8.88} & \multirow{2}{*}{5.98} & 5.55  & 4.47 & \textbf{4.66}  \\
        & & & 9.99 & 4.68 & 6.54 \\
        \midrule
        \multirow{2}{*}{Ours} & \multirow{2}{*}{\textbf{8.15}} & \multirow{2}{*}{\textbf{5.52}} & \textbf{5.38} & \textbf{3.70} & 4.68 \\
        & & & \textbf{9.23} & \textbf{3.76} & \textbf{6.37} \\
        \end{tabular}
    % }
  \caption{Full counting results broken down by class. We report mean average error (MAE), category-averaged masked MAE (CMMAE, see \cref{sec:eval}), and MAE for each class of each attribute. Top row classes: elephant, male, and adult. Bottom row classes: fur, female, and pup. All reported results are the average of 3 runs.}
  \label{tab:results_full}
  \vspace{-5mm}
\end{table}

\vspace{-2mm}
We report our overall results and per-class MAE in \cref{tab:results_full}. Our method shows an 8\% relative improvement over our baseline MAE and CMMAE. The largest improvements come from the sex attribute, where we see a relative improvement of 17\% and 20\% on the male and female classes, respectively. Given the large number of ``unknown'' classifications for this attribute (see \cref{fig:data}A), we hypothesize that this improvement stems from our loss masking approach, which avoids penalizing the network for predicting classes in these abundant unknown regions. We ablate the components of our approach in \cref{tab:results_dens}. We see that the largest single improvements on MAE and CMMAE come from our loss masking technique and the use of cluster-based density maps over the standard fixed kernel approach, respectively.
%, and the largest improvement on CMMAE comes from the use of cluster-based density maps over the standard fixed kernel approach.

In \cref{fig:data}B–C we show qualitative examples of our improvements over the baseline on the age attribute. We notice more accurate predictions both in regions with known classifications (\cref{fig:data}B) as well as unknown regions (\cref{fig:data}C). In the notated examples \#1–\#5, we see that our method is better able to separate small foreground objects from the background, including in regions where classification is difficult, \ie ground truth classifications are unknown. These results are encouraging, however with a mean of 34 objects per image there is still significant room for improvement.

\begin{table}
  \centering
  \small{
      \begin{tabular}{c|c|c|c|c|c|c|c}
        \hline\noalign{\smallskip}
        % Loss & Point & Multi- & & \\
        \multicolumn{3}{c|}{Loss} & Pt & Loss & Mlt & \multirow{2}{*}{MAE} & CM- \\
        $L^c$ & $L^t$ & $L^s$ & Clst & Mask & Task & & MAE  \\
        \noalign{\smallskip}
        \hline\hline
        \noalign{\smallskip}
        \checkmark & & & & & & 8.88 & 5.98 \\
        \checkmark & & & \checkmark & & & 8.83 & 5.79  \\
        \checkmark & \checkmark & & \checkmark & & & 8.68 & 5.72  \\
        \checkmark & \checkmark & & \checkmark & \checkmark & & 8.24 & 5.67  \\
        \checkmark & \checkmark & & \checkmark & \checkmark & \checkmark &  8.29 & 5.59 \\
        \midrule
        \checkmark & \checkmark & \checkmark & \checkmark & \checkmark & \checkmark & \textbf{8.15}  & \textbf{5.52}  \\
        \end{tabular}
    }
  \vspace{-2mm}
  \caption{Ablation study for the components of our approach. We ablate the 6 components described in \cref{sec:gt} and \cref{sec:dens_methods}: multiclass MSE loss ($L^c$), total count MSE loss ($L^t$), segmentation loss ($L^s$), point cluster dot aggregation (``Pt Clst''), loss masking (``Loss Mask''), and multi-task network (``Mlt Task''). We see that each component contributes to an overall 8\% relative improvement in MAE and CMMAE over the baseline (first row vs. final row).}
  \label{tab:results_dens}
  \vspace{-6mm}
\end{table}

\section{Conclusions and future work}
\label{sec:conclusion}
\vspace{-2mm}

We introduce the \datasetname dataset for fine-grained counting using crowd-sourced annotations.
We plan to expand the dataset to include additional locations and seasons, as well as collect a set of expert annotations to allow for a more quantitative study of ground truth generation methods. 

Our initial experimental results are encouraging, indicating an opportunity for computer vision to help scale up research in domains such as animal ecology where crowd-sourced analysis is already taking place. 
In the future we plan to explore detection-based approaches that would enable downstream analysis, \eg behavior study; approaches that make use of the spatiotemporal context provided by time-lapse cameras; and methods for harnessing additional unannotated imagery via unsupervised techniques.

\vspace{-3mm}
\section{Acknowledgements}
\vspace{-2mm}
This work was partially funded by Experiment.com grant ``Understanding Seal Behavior with Artificial Intelligence.'' We thank all who contributed to the funding effort. We also thank Jimmy Freese for project coordination, and Grant Van Horn and Sara Beery for helpful feedback.

%%%%%%%%% REFERENCES
{\small
\bibliographystyle{ieee_fullname}
\bibliography{bib}
}

\end{document}